%% file: main.tex
\documentclass[letterpaper, 10 pt, conference]{ieeeconf}  

\IEEEoverridecommandlockouts                              

\usepackage{times}
\usepackage{epsfig}
\usepackage{graphicx}
\usepackage{amsmath}
\usepackage{bm}
\usepackage{amssymb}

\usepackage{tabularx}
\usepackage{amsmath}
\usepackage{siunitx}
\usepackage{booktabs}
\usepackage{multirow}
\usepackage[table,xcdraw]{xcolor}
\usepackage{float}
\usepackage[normalem]{ulem}
\useunder{\uline}{\ul}{}
\newcommand{\etal}{et al.}
\usepackage{cite}

\usepackage[pagebackref=true,breaklinks=true,colorlinks,bookmarks=false]{hyperref}

\title{\LARGE \bf
Crowdsourced 3D Mapping: A Combined Multi-View Geometry and Self-Supervised Learning Approach 
}

\author{Hemang Chawla, Matti Jukola, Terence Brouns, Elahe Arani, and Bahram Zonooz
\thanks{All authors are with Advanced Research Lab, NavInfo Europe, Eindhoven, The Netherlands.  \tt\small \{hemang.chawla, matti.jukola, terence.brouns, elahe.arani, b.yoosefizonooz\}@navinfo.eu }%
}

\begin{document}

\input{ieee_copyright.tex}
\newpage

\maketitle
\thispagestyle{empty}
\pagestyle{empty}

\input{abstract.tex}

\input{introduction.tex}

\input{related_work.tex}

\input{proposed_framework.tex}
\input{experiments.tex}
\input{conclusion.tex}


\bibliographystyle{IEEEtran}
\bibliography{refbib.bib}



\addtolength{\textheight}{-12cm}   






\end{document}

%% file: ieee_copyright.tex
\onecolumn
{
\noindent
\large
\textbf{This paper has been accepted for publication in the proceedings of 
\textit{IEEE/RSJ International Conference on Intelligent Robots and Systems (IROS)}, Las Vegas, USA, October 25-29, 2020.}

\bigskip\bigskip
\noindent
\large
IEEE Copyright notice:\\\\
\normalsize
\copyright 2020 IEEE. Personal use of this material is permitted. Permission from IEEE must be obtained for all other uses, in any current or future media, including reprinting /republishing this material for advertising or promotional purposes, creating new collective  works,  for  resale  or  redistribution  to  servers  or  lists,  or  reuse  of  any  copyrighted  component  of  this  work  in  other works.

\bigskip\bigskip
\noindent
\large
Cite as:\\\\
\noindent\fbox{%
    \parbox{\textwidth}{%
    \noindent
    \normalsize
       H. Chawla, M. Jukola, T. Brouns, E. Arani, and B. Zonooz, ``Crowdsourced  3D  Mapping:  A  Combined  Multi-View  Geometry  and Self-Supervised  Learning  Approach," \textit{2020 IEEE/RSJ International Conference on Intelligent Robots and Systems (IROS), Las Vegas, USA, IEEE (in press), 2020.}%
    }%
}

\bigskip\bigskip
\noindent
\large
\textsc{Bib}\TeX:\\\\
\noindent\fbox{%
    \parbox{\textwidth}{%
    \noindent
    \normalsize
    \texttt{\noindent @inproceedings\{chawla2020monocular,\\
        author=\{H. \{Chawla\} and M. \{Jukola\} and T. \{Brouns\} and E. \{Arani\} and B. \{Zonooz\}\},\\
        booktitle=\{2020 IEEE/RSJ International Conference on Intelligent Robots and Systems (IROS)\}, \\
        title=\{Crowdsourced  3D  Mapping:  A  Combined  Multi-View  Geometry  and Self-Supervised  Learning  Approach\}, \\
        location=\{Las Vegas, USA\},\\
        publisher=\{IEEE (in press)\},\\
        year=\{2020\}     
        }%
    }%
}
}
\normalsize
\twocolumn

%% file: abstract.tex
\begin{abstract}
The ability to efficiently utilize crowd-sourced visual data carries immense potential for the domains of large scale dynamic mapping and autonomous driving. However, state-of-the-art methods for crowdsourced 3D mapping assume prior knowledge of camera intrinsics. In this work we propose a framework that estimates the 3D positions of semantically meaningful landmarks such as traffic signs without assuming known camera intrinsics, using only monocular color camera and GPS. We utilize multi-view geometry as well as deep learning based self-calibration, depth, and ego-motion estimation for traffic sign positioning, and show that combining their strengths is important for increasing the map coverage. To facilitate research on this task, we construct and make available a KITTI based 3D traffic sign ground truth positioning dataset. Using our proposed framework, we achieve an average single-journey relative and absolute positioning accuracy of \SI[detect-weight=true, detect-family=true, mode=text]{39}{\cm}~ and  \SI[detect-weight=true, detect-family=true, mode=text]{1.26}{\m}~ respectively, on this dataset. 
\end{abstract}

%% file: introduction.tex
\section{Introduction}
\label{section:introduction}

Recent progress in computer vision has enabled the implementation of autonomous vehicle prototypes across urban and highway scenarios \cite{schwarting2018planning}. 
Autonomous vehicles need accurate self-localization in the environment allowing them to plan their actions.
For accuracy in localization, the High Definition (HD) maps of the environments containing information on 3D geometry of road boundaries, lanes, traffic signs, and other semantically meaningful landmarks are necessary.
However, the process of creating these HD maps involves the use of expensive sensors mounted on the collection vehicles \cite{jiao2018machine}, thereby limiting the scale of their coverage.
It is also desired that any changes in the environment, such as the type or positions of traffic signs, are regularly reflected in the map. Therefore, the creation and maintenance of HD maps at scale remain a challenge.

\begin{figure}[t]
\begin{center}
   \includegraphics[width=1\linewidth]{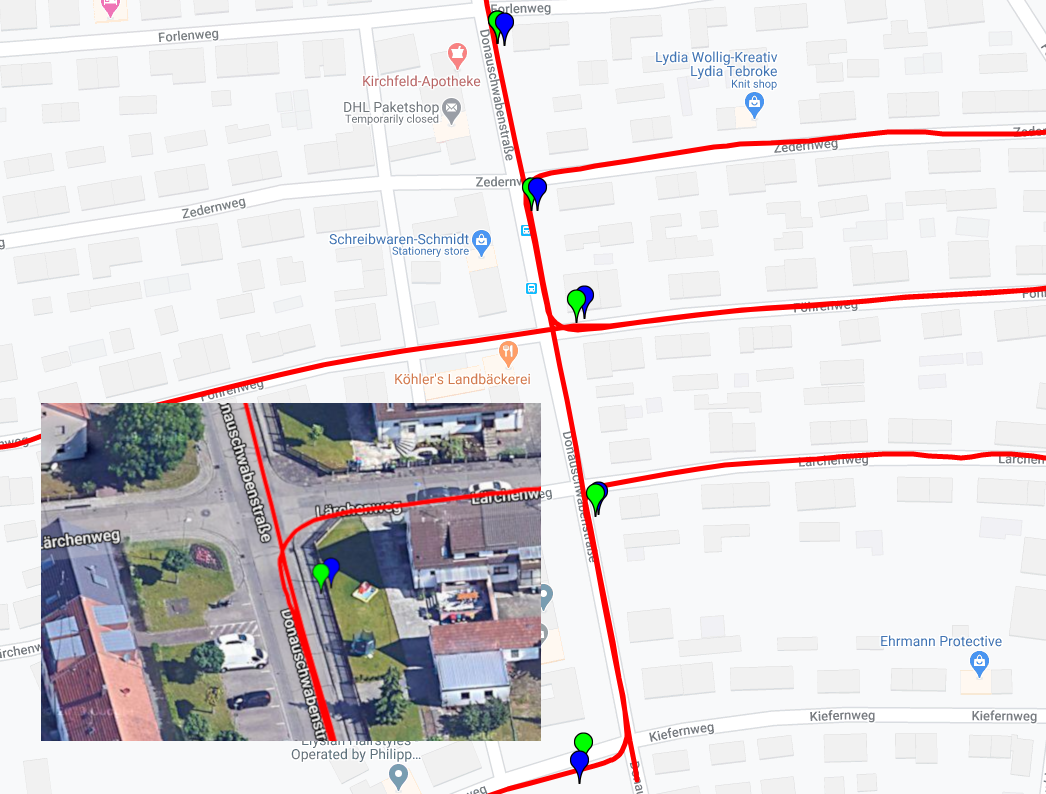}
\end{center}
   \caption{3D traffic sign triangulation for KITTI. The estimated signs are shown in blue, and the corresponding ground truth signs are shown in green. The computed path of the vehicle used for triangulation is depicted in red.}
\label{fig:self_calib_sensitivity_oat}
\end{figure}

To extend the map coverage to more regions or update the landmarks over time, crowdsourced maps are an attractive solution.
However, in contrast with the automotive data collection vehicles with high grade calibrated sensors, crowdsourced maps would require the use of consumer-grade sensors whose intrinsics may be unknown, or change over time.
The commonly available sensors for crowdsourced mapping are a monocular color camera and a global positioning system (GPS). To utilize these sensors for crowdsourced mapping, it is required to perform camera self-calibration followed by monocular depth or ego-motion estimation.
Over the years, geometry as well as deep learning based approaches have been proposed to compute the camera intrinsics \cite{bogdan2018deepcalib,gordon2019depthwild, schonberger2016structure}, and estimate the depth/ego-motion \cite{mur2015orb, engel2017direct, gordon2019depthwild, zhou2017unsupervised} from a sequence of images. However, the state-of-the-art solution to crowdsourced mapping assumes the camera intrinsics to be known a priori, and relies upon only geometry based ego-motion estimation \cite{dabeer2017end}.

Geometry based approaches for self-calibration and visual depth/ego-motion estimation often depend on carefully designed features and matching them across frames. Thus, they fail in scenarios with limited features such as highways, during illumination change, occlusions, or have poor matching due to structure repetitiveness. 
Recently, deep learning based approaches for camera self-calibration as well as depth and ego-motion estimation have been proposed \cite{zhou2018deeptam, zhou2017unsupervised, godard2018digging, gordon2019depthwild}. These methods perform in an end-to-end fashion and often being self-supervised, enable application in challenging scenarios. They are usually more accurate than geometry based approaches on short linear trajectories, resulting in a higher local agreement with the ground truth \cite{zhou2017unsupervised, gordon2019depthwild}. 
Moreover, deep learning based approaches can estimate monocular depth from a single frame as opposed to geometry based approaches that require multiple frames.
Nonetheless, the localization accuracy of geometry based approaches is higher for longer trajectories due to loop closure and bundle adjustment.
Therefore, we hypothesize that eliminating the requirement to know the camera intrinsics a priori while mapping through a hybrid of geometry and deep learning methods, will increase the global map coverage and enhance the scope of its application.

In this work, we focus on the 3D positioning of traffic signs, as it is critical to the safe performance of autonomous vehicles,
and is useful for traffic inventory and sign maintenance. We propose a framework for crowdsourced 3D traffic sign positioning that combines the strengths of geometry and deep learning approaches to self-calibration and depth/ego-motion estimation. 
Our contributions are as follows:
\begin{itemize}
    \item We evaluate the sensitivity of the 3D position triangulation to the accuracy of the self-calibration.
    \item We quantitatively compare deep learning and multi-view geometry based approaches to camera self-calibration, as well as depth and ego-motion estimation for crowdsourced traffic sign positioning.
    \item We demonstrate crowdsourced 3D traffic sign positioning using only GPS information and a monocular color camera without the prior knowledge of camera parameters.
    \item We show that combining the strengths of deep learning with multi-view geometry is important for increased map coverage.
    \item To facilitate evaluation and comparison on this task, we construct and provide an open source 3D traffic sign ground truth positioning dataset on KITTI\footnote{\url{https://github.com/hemangchawla/3d-groundtruth-traffic-sign-positions.git}}.
\end{itemize}

%% file: related_work.tex
\section{Related Work}
\label{section:related_work}
\paragraph{Traffic sign 3D positioning}
Arnoul \etal~\cite{arnoul1996traffic} used a Kalman filter for tracking and estimating positions of traffic signs in static scenes.
In contrast, Madeira \etal~\cite{madeira2005automatic} estimated traffic sign positions through least-squares triangulation 
using GPS, Inertial Measurement Unit (IMU), and wheel odometry. 
Approaches using only a monocular color camera and GPS were also proposed \cite{krsak2011traffic, welzel2014accurate}. However, Welzel~\etal \cite{welzel2014accurate} utilized prior information about the size and height of traffic signs to achieve an average absolute positioning accuracy up to 1m. 
A similar problem of mapping the 3D positions and orientations of traffic lights 
was tackled by Fairfield \etal~\cite{fairfield2011traffic}.
For related tasks of 3D object positioning and distance estimation, deep learning approaches \cite {chen2016monocular,ku2019monocular,qin2019monogrnet, zhu2019learning} have been proposed. However, they primarily focus on volumetric objects, ignoring the near-planar traffic signs.
Recently, Dabeer \etal~\cite{dabeer2017end} proposed an approach to crowdsource the 3D positions and orientations of traffic signs using cost-effective sensors with known camera intrinsics, and achieved a single journey average relative and absolute positioning accuracy of 46 cm and 57 cm respectively.
All the above methods either relied upon
collection hardware dedicated to mapping the positions of traffic control devices or assumed known accurate camera intrinsics.
\paragraph{Camera self-calibration}
Geometry based approaches for self-calibration use two or more views of the scene to estimate the 
focal lengths 
\cite{bocquillon2007constant,gherardi2010practical}, 
while often fixing the principal point at the image center \cite{de1998self}.
Structure from motion (SfM) reconstruction using a sequence of images has also been applied 
for self-calibration 
\cite{pollefeys2008detailed, schonberger2016structure}. 
Moreover, deep learning approaches have been proposed to estimate the camera intrinsics using a single image through direct supervision \cite{lopez2019deep, rong2016radial,workman2015deepfocal,zhuang2019degeneracy}, or as part of a multi-task network \cite{bogdan2018deepcalib,gordon2019depthwild}. While self-calibration is essential for crowdsourced 3D traffic sign positioning, its utility has not been evaluated until now. 

\begin{figure*}[thbp!]
\begin{center}
    \includegraphics[width=0.88\linewidth]{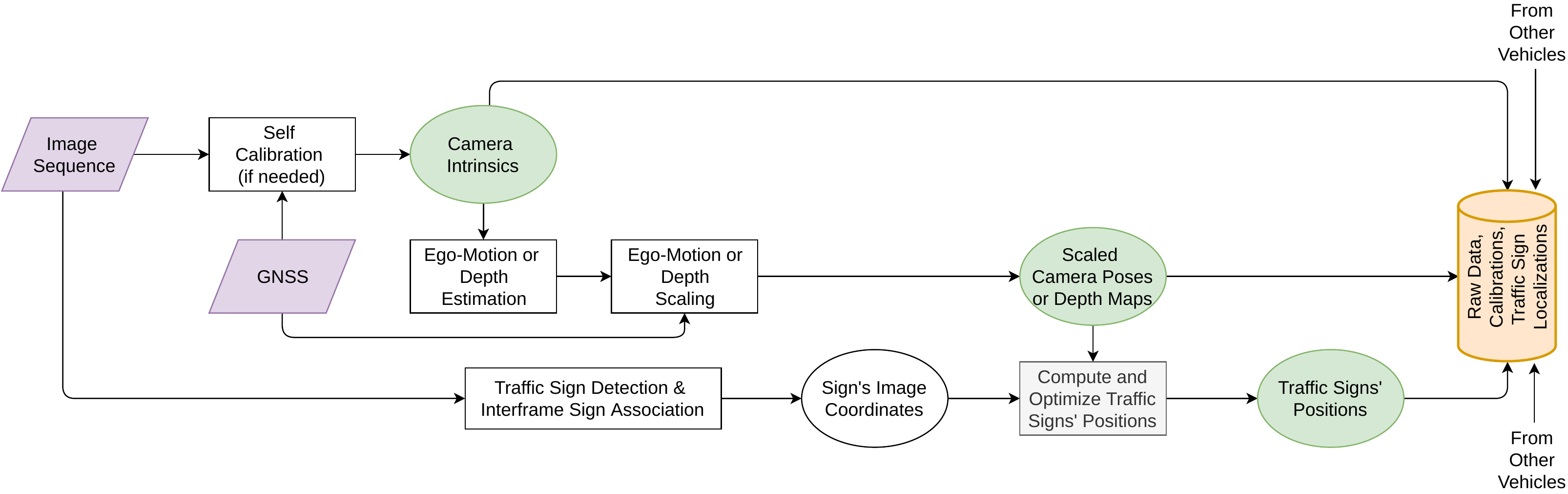}
\end{center}
   \caption{Single Journey 3D Traffic Sign Positioning Framework. Components in purple represent the inputs to the framework. Components in green represent the outputs of the three primary steps of the system. The crowdsourced mapping engine in orange depicts the traffic sign positioning data coming from different vehicles. Traffic Sign Positioning can be performed using ego-motion or depth estimation. Fig. \ref{fig:TwoApproaches} elaborates upon the respective approaches.}
\label{fig:SingleJourney}
\end{figure*}

\paragraph{Monocular Depth and Ego-Motion estimation}
Multi-view geometry based monocular visual odometry (VO), and simultaneous localization and mapping (SLAM) estimate the camera trajectory using visual feature matching and local bundle adjustment \cite{klein2007parallel, mur2015orb}, or through minimization of the photometric reprojection error \cite{engel2014lsd, engel2017direct, newcombe2011dtam}.
Supervised learning approaches predict monocular depth \cite{cao2017estimating, eigen2014depth, liu2015learning} and ego-motion \cite{wang2017deepvo, zhou2018deeptam} using ground truth depths and trajectories, respectively. In contrast, self-supervised approaches jointly predict ego-motion and depth utilizing image reconstruction as a supervisory signal \cite{casser2019unsupervised1, godard2018digging, gordon2019depthwild, zhou2017unsupervised, godard2017unsupervised, li2018undeepvo, zhan2018unsupervised}.
Self-supervised depth prediction has also been integrated with geometry based direct sparse odometry \cite{engel2017direct} as a virtual depth signal \cite{yang2018deep}.
However, some of these self-supervised approaches rely upon stereo image pairs during training \cite{yang2018deep, godard2017unsupervised, li2018undeepvo, zhan2018unsupervised}.

%% file: proposed_framework.tex
\section{Method}
\label{section:system_overview}
In this section, we describe our proposed system for 3D traffic sign positioning. The input is a sequence of $n$ color images $\mathbf{I} = \{I_0, \dots, I_{n-1} \}$ of width $w$ and height $h$, and corresponding GPS coordinates $\mathbf{G} = \{g_0, \dots, g_{n-1} \}$. The output is a list of $m$ detected traffic signs with the corresponding class identifiers $C_i$, absolute positions $p^{abs}_i$, and the relative positions $\{p^{rel}_{i,j}\}$ with respect to the corresponding frames $j$ in which the sign was detected.
An overview of the proposed system for 3D traffic sign positioning is depicted in Fig. \ref{fig:SingleJourney}.
Our system comprises of the following key modules:

\subsection{Traffic Sign Detection \& Inter-frame Sign Association}
\label{section:sign_detection}
The first requirement for the estimation of 3D positions of traffic signs is detecting their coordinates in the image sequence and identifying their class.
The output of this step is a list of 2D bounding boxes enclosing the detected signs, and their corresponding track and frame numbers. Using the center of the bounding box we extract the coordinates of the traffic sign in the image. However, we disregard those bounding boxes that are detected at the edge of the images to account for possible occlusions. 
\subsection{Camera Self-Calibration} 
\label{section:self_calibration}
For utilizing the crowdsourced image sequences to estimate the 3D positions of traffic signs, we must perform self-calibration for cameras whose intrinsics are previously unknown. For this work, we utilize the pinhole camera model. From the set of geometry based approaches, we evaluate the Structure from Motion based method using Colmap \cite{schonberger2016structure}.
Note that self-calibration suffers from ambiguity for the case of forward motion with parallel optical axes \cite{bocquillon2007constant}.
Therefore we only utilize those parts of the sequences in which the car is turning. To extract the sub-sequences in which the car is turning, the Ramer-Douglas-Peucker (RDP) algorithm \cite{ramer1972iterative, douglas1973algorithms} is used.
From the deep learning based approaches, we evaluate the Self-Supervised Depth From Videos in the Wild (VITW) \cite{gordon2019depthwild}.
The burden of annotating training data \cite{lopez2019deep, zhuang2019degeneracy} makes supervised approaches inapplicable to crowdsourced use-cases.

\begin{figure}[b!]
\begin{center}
  \includegraphics[width=1\linewidth]{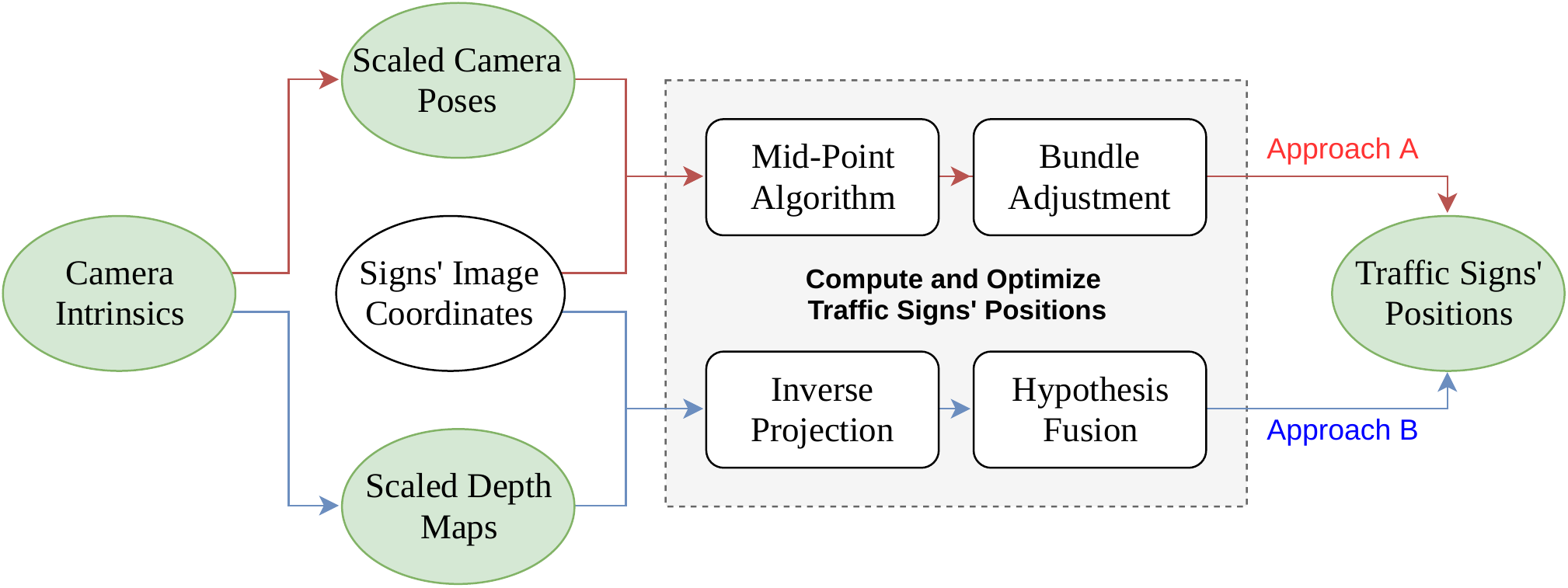}
\end{center}
\caption{Two approaches to estimating 3D traffic sign positions. Approach A uses camera intrinsics to perform ego-motion estimation and generate camera poses. Approach B uses camera intrinsics to perform depth estimation and generate a depth map per image. Approaches A and B utilize the camera poses and depth maps along with the traffic signs' image coordinates to compute the 3D positions of traffic signs. The modules in green depict the primary steps from Fig. \ref{fig:SingleJourney}.}
\label{fig:TwoApproaches}
\end{figure}

\subsection{Camera Ego-Motion and Depth Estimation} 
\label{section:ego_motion_and_depth}
Given the camera calibration, the 3D traffic sign positioning requires the computation of the camera ego-motion or depth as shown in Fig. \ref{fig:SingleJourney} and \ref{fig:TwoApproaches}.

\paragraph{Ego-Motion} For applying approach A described in Fig. \ref{fig:TwoApproaches} to 3D traffic sign positioning, the ego-motion of the camera must be computed from the image sequence. Note that camera-calibration through Colmap involves SfM, but only utilizes those sub-sequences which contain a turn (Sec. \ref{section:self_calibration}). Therefore, we evaluate state-of-the-art geometry based monocular approach ORB-SLAM \cite{mur2015orb} against self-supervised Monodepth 2 \cite{godard2018digging} and VITW. While the geometry based approaches compute the complete trajectory for the sequence, the self-supervised learning based approaches output the camera rotation and translation per image pair. The adjacent pair transformations are then concatenated to compute the complete trajectory. 
After performing visual ego-motion estimation, we use the GPS coordinates to scale the estimated trajectory. First, we transform the GPS geodetic coordinates to local East-North-Up (ENU) coordinates. 
Thereafter, using the Umeyama's algorithm \cite{umeyama1991least}, a similarity transformation, (rotation $R_e$, translation $t_e$, and scale $s_e$) is computed that scales and aligns the estimated camera positions ($t_{0,j},  \forall j=1\dots n$) with the ENU positions ($g_{0,j}, \forall j=1\dots n$) minimizing the mean squared error 
between them.
The scaled and aligned camera positions are therefore given by
\begin{equation}
\label{eq:scaled_camera_positions}
   t'_{0,j} = s_eR_et_{0,j} + t_e.
\end{equation}
Thereafter, this camera trajectory is used for computation of the 3D traffic sign positions as described in section \ref{section:sign_positioning}.
 
\begin{figure}[b!]
\begin{center}
    \includegraphics[width=0.65\linewidth]{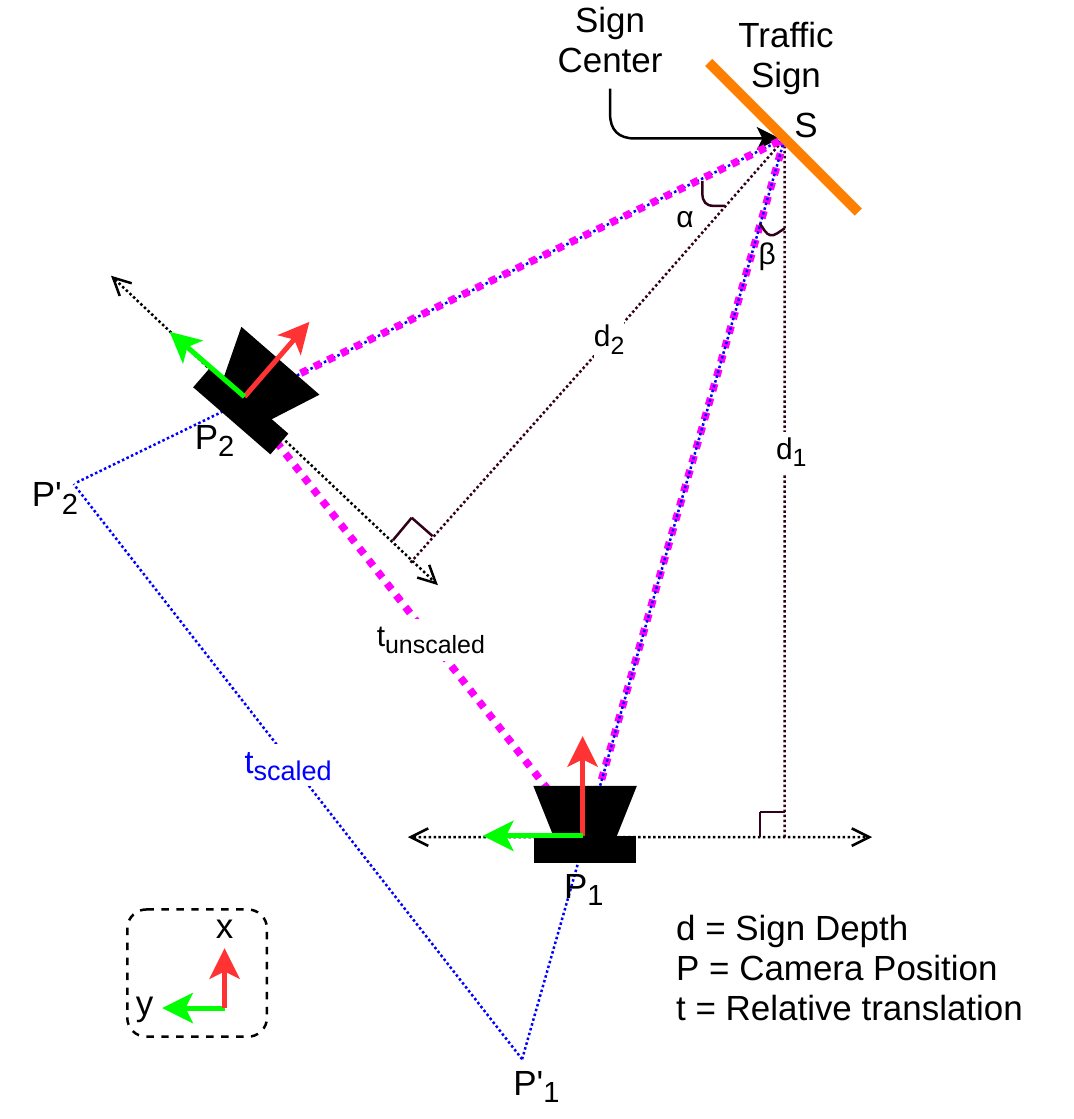}
\end{center}
   \caption{Scaling of depth using similar triangles (top view). For any object S (e.g. traffic sign) in the scene, the depths d\textsubscript{1} and d\textsubscript{2} from consecutive camera positions P\textsubscript{1} and P\textsubscript{2} can be scaled with a common factor $s_d$. Since $\triangle$ SP\textsubscript{1}P\textsubscript{2} $\sim \triangle$ SP'\textsubscript{1}P'\textsubscript{2}, this scaling does not affect angles $\alpha$ and $\beta$. The scaling factor $s_d$ is given by the ratio of t\textsubscript{scaled} to t\textsubscript{unscaled}, where t represents the relative translation between the frames.}
\label{fig:DepthScalingTriangles}
\end{figure}

\paragraph{Monocular Depth}
For applying approach B described in Fig. \ref{fig:TwoApproaches} to 3D traffic sign positioning, dense monocular depth maps are needed. To generate the depth maps, we evaluate the self-supervised approaches, Monodepth 2, and VITW. These approaches simultaneously predict the monocular depth as well as the ego-motion of the camera.
While the estimated dense depth maps maintain the relative depth of the observed objects, we obtain metric depth by preserving forward and backward scale consistency.
Given camera calibration matrix $K$, the shift in pixel coordinates due to rotation $R_{j+1,j}$ and translation $t_{j+1,j}$ between adjacent frames $j$ and $j+1$, is given by
\begin{equation}
\label{eq:pixel_shift}
    d(c_{j+1})c_{j+1} = KR_{j+1,j}K^{-1}d(c_j)c_j + Kt_{j+1,j},
\end{equation}
where $d(c_j)$ and $d(c_{j+1})$ represent the unscaled depths corresponding to the homogeneous coordinates of pixels $c_j$ and $c_{j+1}$. 
By multiplying equation \ref{eq:pixel_shift} with forward scale estimate ${s'_d}_{j+1,j}$, it is seen that scaling the relative translation $t_{j+1,j}$ similarly scales the depths $d(c_j)$ and $d(c_{j+1})$. This is also explained through the concept of similar triangles in Fig. \ref{fig:DepthScalingTriangles}.
Given relative ENU translation $g_{j+1,j}$, we note that the scaled relative translation is given by,
\begin{equation}
    {s'_d}_{j+1,j} \cdot t_{j+1,j} =  g_{j+1,j}.
\end{equation}
Therefore, the forward scale estimate
\begin{equation}
 \label{eq:scale_depth}
    {s'_d}_{j+1,j} = \frac{\lVert g_{j+1,j} \rVert}{\lVert t_{j+1,j} \rVert}.
\end{equation}
Similarly the backward scale estimate ${s'_d}_{j, j-1}$ is computed. 
Accordingly, for frames $j = 1 \cdots n-2$, the scaling factor ${s_d}_j$ is given by the average of forward and backward scale estimates, ${s'_d}_{j+1,j}$ and ${s'_d}_{j, j-1}$.
Thereafter, these scaled dense depth maps are used for computation of the 3D traffic sign positions as described in section \ref{section:sign_positioning}.

\subsection{3D Positioning and Optimization} 
\label{section:sign_positioning}
For the final step of estimating and optimizing the 3D positions of the detected traffic signs, we adopt two approaches as shown in Fig. \ref{fig:TwoApproaches}. 

\paragraph{Approach A}
In this approach, the estimated camera parameters, the computed and scaled ego-motion trajectory, and the 2D sign observations in images are used to compute the sign position through triangulation. 
For a sign $S_i$ observed in $k$ frames, we compute the initial sign position estimate $p_i^{init}$ using the mid-point algorithm \cite{szeliski2010computer}. 
Thereafter, non-linear Bundle Adjustment (BA) is applied to refine the initial estimate by minimizing the reprojection error to output
\begin{equation}
\label{eq:ba}
    p_i^{abs} = \underset{p_i}{\text {arg min}} \left(\sum_j\lVert K (R_{j,0}p_i + t_{j,0}) - c_{i,j}\rVert ^2\right).
\end{equation}
To compute the sign positions $p_{i,j}^{rel}$ relative to frames $j$, the estimated absolute sign position $p_i^{abs}$ is projected to the corresponding frames in which it was observed 
\begin{equation}
\label{eq:rel_traingulation_positioning}
    p_{i,j}^{rel} = R_{j,0}p_i^{abs} + t_{j,0}.
\end{equation}
If the relative depth of the sign is found to be negative, triangulation of that sign is considered to be failed. We can use this approach with the full trajectory of the sequence or with short sub-sequences corresponding to the detection tracks. The use of full and short trajectory for triangulation is compared in section \ref{sec:3d_analysis}. 

\paragraph{Approach B}
In approach B, the estimated camera parameters, the scaled dense depth maps, and the 2D sign observations in images  are used to compute the 3D traffic sign positions through inverse projections. For a sign $S_i$ observed in $k$ frames, each corresponding depth map produces a sign position hypothesis given by
\begin{equation}
\label{eq:relative_depth_positioning}
    p_{i,j}^{rel} = {s_d}_j \cdot d(c_{i,j}) K^{-1}c_{i,j}
\end{equation}
where $c_{i,j}$ represents the pixel coordinate of sign $i$ in the frame $j$, and ${s_d}_j$ is the corresponding depth scaling factor. Since, sign depth estimation may not be as reliable beyond a certain distance, we discard that sign position hypotheses whose estimated relative depth is more than 20m. 
For computing the absolute coordinates of the sign, each relative sign position is projected to the world coordinates, and their centroid is computed as the absolute sign position,
\begin{equation}
\label{eq:abs_depth_positioning}
    p_i^{abs} = \dfrac{\sum_{j}[R_{0,j} | t_{0,j}] p_{i,j}^{rel}}{k}.
\end{equation}

Finally, for both the above approaches, the metric absolute positions of traffic signs are converted back to the GPS geodetic coordinates.
 

%% file: experiments.tex
\section{Experiments}
In order to evaluate the best approach to 3D traffic sign positioning, it is pertinent to consider the impact of the different components on the overall accuracy of the estimation.
First, we analyze the sensitivity of 3D traffic sign positioning performance against the camera calibration accuracy, demonstrating the importance of good self-calibration. Thereafter, we compare approaches to ego-motion and depth estimation, and camera self-calibration that compose the 3D sign positioning system. Finally, the relative and absolute traffic sign positioning errors corresponding to the approaches A and B are evaluated. For the above comparisons, we use the traffic signs found in the raw KITTI odometry dataset \cite{geiger2012we}, sequences (Seq) 0 to 10 (Seq 3 is missing from the raw dataset), unless specified otherwise.

\subsection{Ground Truth Traffic Sign Positions}
While 3D object localization datasets usually contain annotations for volumetric objects, such as vehicles and pedestrians, such annotations for near-planar objects like traffic signs are lacking. Furthermore, related works dealing with 3D traffic sign positioning have relied upon closed source datasets \cite{dabeer2017end, welzel2014accurate}. Therefore we generate the ground truth (GT) traffic sign positions required for validation of the proposed approaches in the KITTI dataset. We choose the challenging KITTI dataset, commonly used for benchmarking ego-motion, as well as depth estimation because it contains the camera calibration parameters, and synced LiDAR information that allows annotation of GT 3D traffic sign positions. 

\begin{figure}[!htbp]
\begin{center}
    \includegraphics[width=0.9\linewidth]{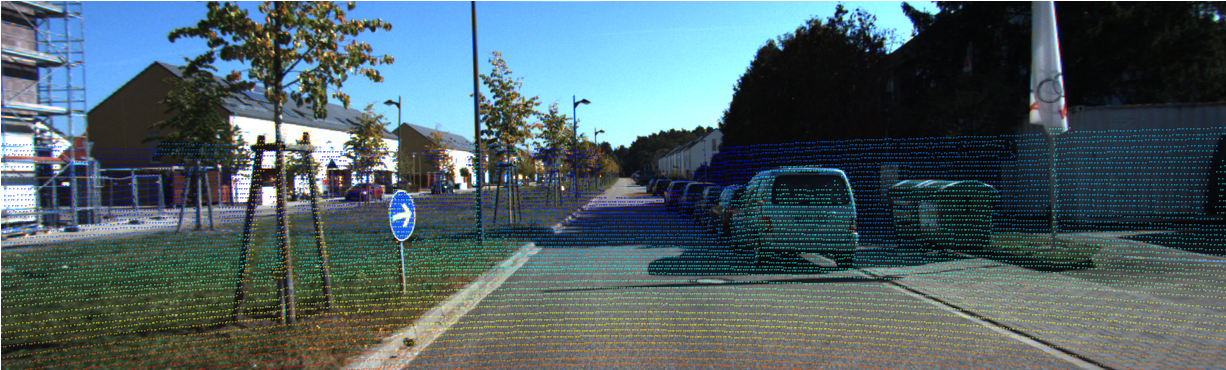}
\end{center}
  \caption{Ground truth traffic sign positions are annotated using images and mapped LiDAR scans.}\label{fig:anno_gt}
\end{figure}
As shown in Fig. \ref{fig:anno_gt}, the LiDAR scans corresponding to the images captured, along with the GT trajectory poses are used to annotate the absolute as well as relative GT positions of the traffic signs. In total, we have annotated 73 signs across the 10 validation sequences.

\subsection{Sensitivity to Camera Calibration}
The state-of-the-art approach to 3D sign positioning relies upon multi-view geometry triangulation.
In this section, we analyze the sensitivity of this method to the error in the estimate of camera focal lengths and principal point. To evaluate the sensitivity, we introduce error in the GT camera intrinsics and perform SLAM, both with (w/) and without (w/o) loop closure (LC) using the incorrect camera matrix, followed by the sign position triangulation using the full trajectory. Its performance for the corresponding set of camera intrinsics is then evaluated as the mean of relative positioning error normalized by the number of signs successfully triangulated. We perform this analysis for KITTI Seq 5 (containing multiple loops) and 7 (containing a single loop). For each combination of camera parameters, we repeat the experiment 10 times and report the minimum of the above metric.

\begin{figure}[!htbp]
\begin{center}
    \includegraphics[width=0.7\linewidth]{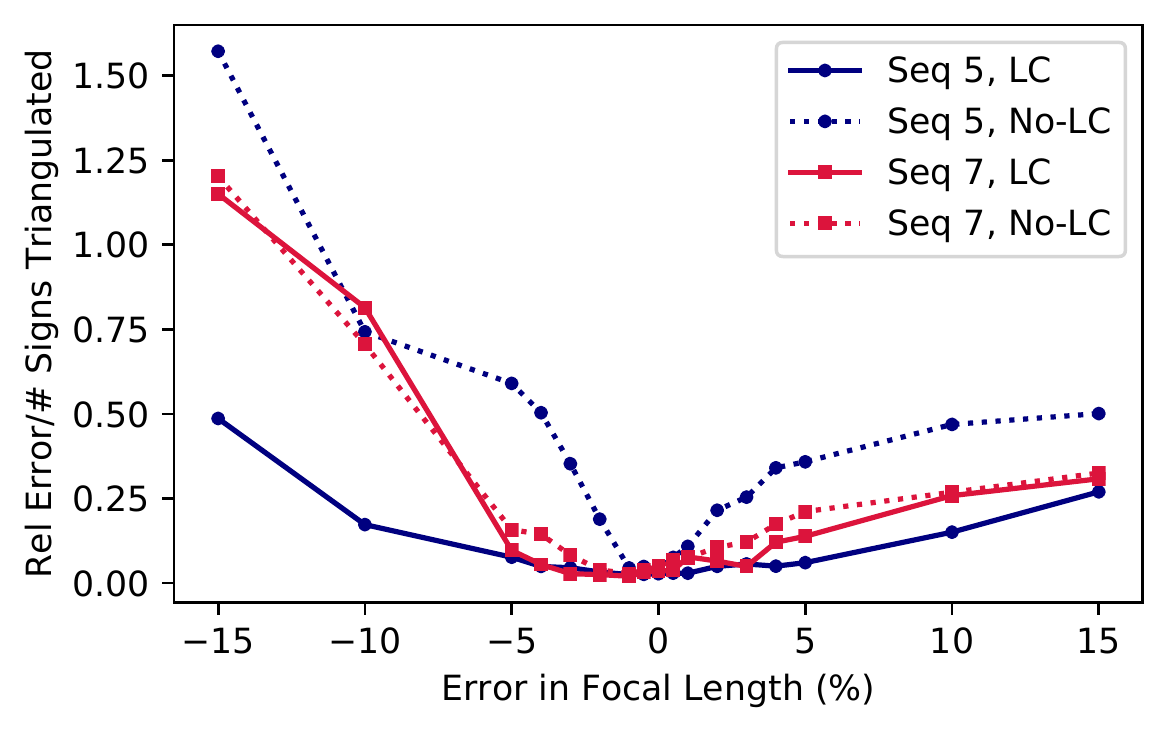}\\
    \includegraphics[width=0.7\linewidth]{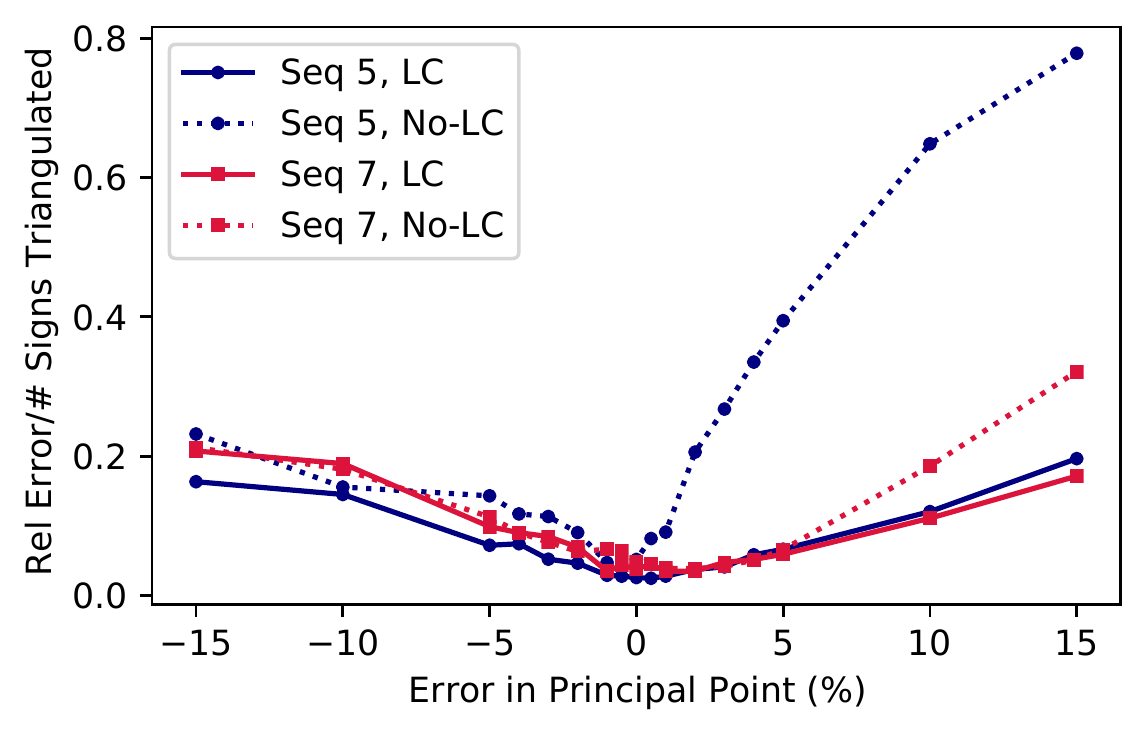}
\end{center}
  \caption{OAT sensitivity analysis. Top: Sensitivity analysis for the error in focal lengths. Bottom: Sensitivity analysis for the error in principal point.}
  \label{fig:OAT_sensitivity}
\end{figure}

\paragraph{One-at-a-Time}
The one-at-a-time (OAT) sensitivity analysis measures the effect of error (-15\% to +15\%) in a single camera parameter while keeping the others at their GT values. Fig. \ref{fig:OAT_sensitivity} shows the sensitivity of sign positioning performance to the error in focal lengths ($f_x$ and $f_y$ are varied simultaneously) and principal point ($c_x$ and $c_y$ are varied simultaneously). The performance w/ LC is better than that w/o LC. Furthermore, the performance gap between triangulation w/ and w/o LC is higher with a higher number of loops (Seq 5). Moreover, the triangulation is more sensitive to underestimating the focal length, and overestimating the principal point, primarily at large errors.

\begin{figure}[tb!]
\begin{center}
    \includegraphics[width=1\linewidth]{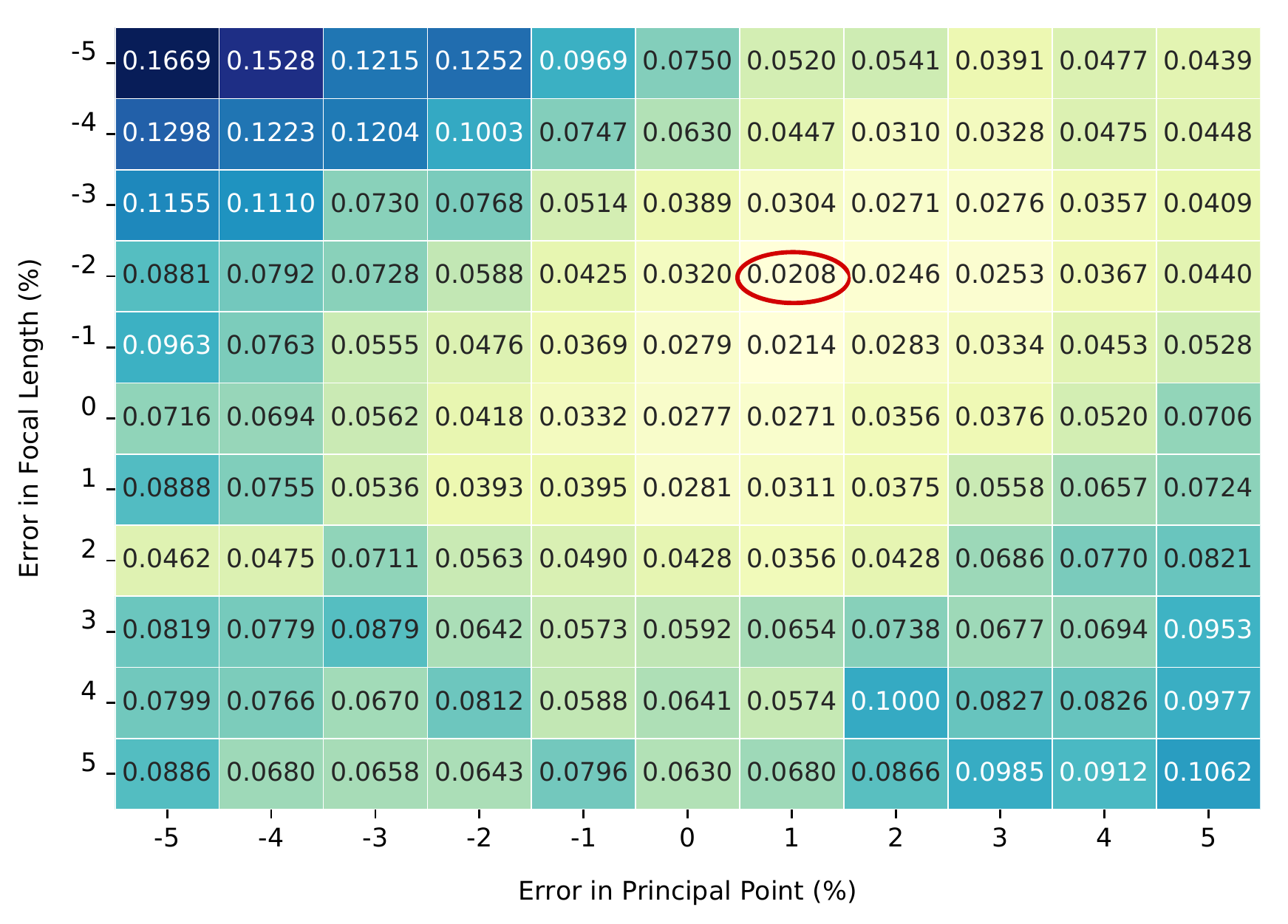}\\
    \includegraphics[width=1\linewidth]{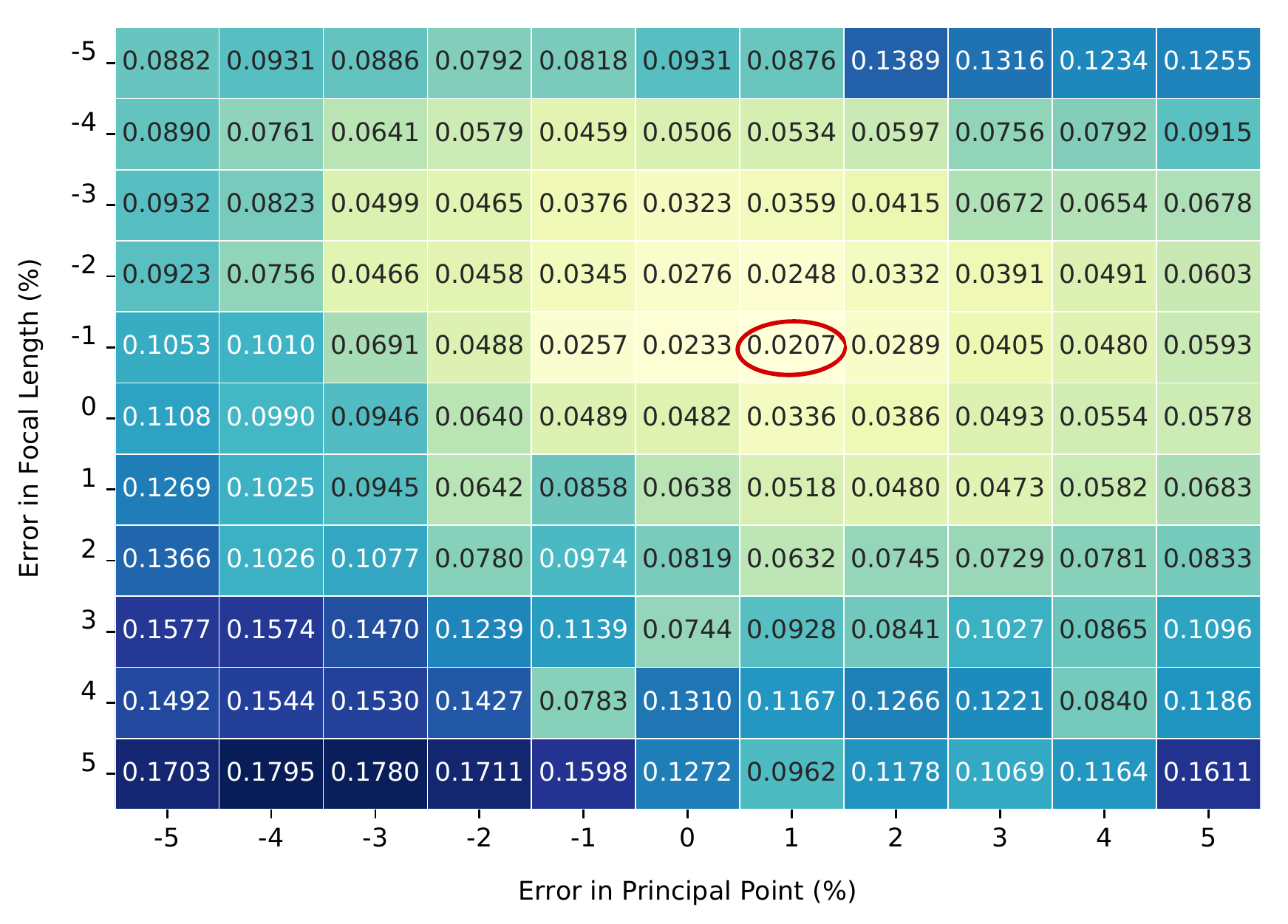}
\end{center}
  \caption{Interaction sensitivity analysis for focal lengths and principal point. Top: Sensitivity of Seq 5 with multiple loop closures. Bottom: Sensitivity of Seq 7 with single loop closure. The best performance is marked with a red oval.}
  \label{fig:int_sensitivity}
\end{figure}

\paragraph{Interaction}
The interaction sensitivity analysis measures the effect of error (-5\% to +5\%) while varying the focal lengths and the principal point simultaneously. Fig. \ref{fig:int_sensitivity} shows the sensitivity to the combined errors in focal lengths and principal point for Seq 5 and Seq 7. The sensitivity to varying the principal point is higher than the sensitivity to varying the focal length for both the sequences. Furthermore for this shorter range of errors, underestimating the focal length and overestimating the principal point results in a better performance than contrariwise. 
This is in contrast to the observed effect when the percentage errors in intrinsics are higher (cf. Fig. \ref{fig:OAT_sensitivity}). Note that the best performance is not achieved at zero percentage errors for the focal length and principal point. We conclude that accurate estimation of the camera intrinsics is pertinent for accurate sign positioning.

\subsection{Sign Positioning Components Analysis}
In order to compute the sign positions, we need the camera intrinsics through self-calibration, and the ego-motion/depth maps as shown in Fig. \ref{fig:SingleJourney}. Here we quantitatively compare state-of-the-art deep learning and multi-view geometry based methods to monocular camera self-calibration, as well as depth and ego-motion estimation. For these experiments, Monodepth 2 and VITW are trained on 44 sequences from KITTI raw in the \textit{city}, \textit{residential}, and \textit{road} categories.

\paragraph{Self-Calibration}
Table \ref{table:self-calibration} shows the average percentage error for self-calibration with VITW and Colmap. VITW estimates the camera intrinsics for each pair of images in a sequence. Therefore, we compute the mean ($\mu$) and median (m) of each parameter across image pairs as the final estimate. To evaluate the impact of the turning radius on self-calibration with VITW, we also compute the parameters considering only those frames detected as part of a turn (through the RDP algorithm). Multi-view geometry based Colmap gives the lowest average percentage error for each parameter. The second best self-calibration estimation is given by VITW Turns (m). However, both of the above fail in self-calibrating the camera using Seq 4, which does not have any turns. For such a sequence, VITW (m) performs better than VITW ($\mu$). All methods underestimate the focal length, and overestimate the principal point. Moreover, VITW estimates the focal length with higher magnitude of error compared to that of the principal point. 
The upper bound for error estimate of $f_x$ is inversely proportional to the amount of rotation  $r_y$ about the $y$ axis \cite{gordon2019depthwild}. . 
Therefore, estimates of $f_x$ and $c_x$ are better than that of $f_y$ and $c_y$ for all methods, because of the near-planar motion in the sequences.

\begin{table}[tb!]
\caption{Comparison of Self-Calibration Approaches.}
\label{table:self-calibration}
\centering
\resizebox{\columnwidth}{!}{%
\begin{tabular}{|l|cccc|}
\hline
\textbf{Method} & \bm{$\delta f_x$\%} & \bm{$\delta f_y$\%} & \bm{$\delta c_x$\%} & \bm{$\delta c_y$\%} \\ \hline
\textbf{Colmap} & {\bf -0.90}$\pm$1.51 & {\bf -0.90}$\pm$1.51 & {\bf 0.77}$\pm$0.40 & {\bf 1.76}$\pm$0.34 \\
\textbf{VITW ($\mu$)} & -23.12$\pm$6.68 & -25.55$\pm$3.28 & 1.07$\pm$0.59 & 3.64$\pm$1.14 \\
\textbf{VITW (m)} & -23.20$\pm$7.41 & -25.33$\pm$3.63 & 0.99$\pm$0.61 & 3.32$\pm$1.10 \\
\textbf{VITW Turns ($\mu$)} & -14.69$\pm$4.34 & -22.96$\pm$2.83 & 1.25$\pm$0.38 & 4.08$\pm$1.17 \\
\textbf{VITW Turns (m)} & -11.82$\pm$6.65 & -22.62$\pm$2.92 & 1.20$\pm$0.34 & 3.88$\pm$1.18\\ \hline
\end{tabular}%
}
\end{table}

\paragraph{Ego-Motion}
Table \ref{table:ego_motion} shows the average absolute trajectory errors (ATE) in meters for full \cite{horn1987closed} and 5-frame sub-sequences (ATE-5) \cite{zhou2017unsupervised} from ego-motion estimation.
The multi-view geometry based ORB-SLAM w/ LC has the lowest ATE full. However, ORB-SLAM w/o LC has a higher local agreement with the GT trajectory depicted by the lowest ATE-5 mean of \SI[detect-weight=true, detect-family=true, mode=text]{0.0137}{\m}.
Both ORB-SLAM methods suffer from track failure for Seq 1, unlike Monodepth 2 and VITW. For Seq 1, VITW has a better performance than Monodepth 2.
While Monodepth 2 has the lowest ATE-5 Std, and an ATE-5 mean similar to that of ORB-SLAM w/o LC,
its ATE full is much higher than that of the ORB-SLAM methods.  

\begin{table}[!t]
\caption{Comparison of Ego-Motion Estimation Approaches.}
\label{table:ego_motion}
\centering
\resizebox{\columnwidth}{!}{%
\begin{tabular}{|l|ccc|}
\hline 
\textbf{Method}  & \textbf{ATE Full} & \textbf{ATE-5 Mean} & \textbf{ATE-5 Std} \\ \hline 
\textbf{ORB-SLAM (w/ LC)}    & {\bf 17.034} & 0.015 & 0.017 \\
\textbf{ORB-SLAM (w/o LC)}   & 37.631       & {\bf 0.014} & 0.015 \\ 
\textbf{VITW (Learned)}      & 85.478       & 0.031       & 0.026 \\ 
\textbf{Monodepth2 (Average)} & 66.494      & 0.014       & {\bf 0.010} \\ \hline
\end{tabular}%
}
\end{table}

\paragraph{Depth}
Table \ref{table:depth} shows the performance of depth estimation based on the metrics defined by Zhou \etal \cite{zhou2017unsupervised}. While Monodepth 2 outperforms VITW in all the metrics, its training uses the average camera parameters from the dataset being trained on, thereby necessitating some prior knowledge about the dataset.  
\begin{table*}[!htbp]
\caption{Comparison of Depth Estimation Approaches.}
\label{table:depth}
\centering
\begin{tabular}{|l|ccccccc|}
\hline
\textbf{Method} & $\downarrow$\textbf{Abs Rel Diff} & $\downarrow$\textbf{Sq Rel Diff} & $\downarrow$\textbf{RMSE} & $\downarrow$\textbf{RMSE (log)} & $\uparrow$\bm{$\delta < 1.25$} & $\uparrow$\bm{$\delta < 1.25^2$} & $\uparrow$\bm{$\delta < 1.25^3$} \\ \hline
\textbf{VITW}       & 0.172 & 1.325 & 5.662 & 0.246 & 0.767 & 0.920 & 0.970 \\ 
\textbf{Monodepth2} & \textbf{0.138} & \textbf{1.132} & \textbf{5.121} & \textbf{0.211} & \textbf{0.838} & \textbf{0.948} & \textbf{0.979} \\ \hline
\end{tabular}
\end{table*}

\begin{table*}[!bhtp]
\caption{Average Relative Sign Positioning Errors for Approaches A and B. Best result is in bold. The second best is underlined.}
\label{tab:kitti_rel_final}
\centering
\begin{tabular}{|l|lllll|lllll||lll|lll|}
\hline
\multirow{2}{*}{} & \multicolumn{10}{c||}{\textbf{Approach A}} & \multicolumn{6}{c|}{\textbf{Approach B}} \\ \cline{2-17} 
 & \multicolumn{5}{c|}{\textbf{ORB-SLAM w\textbackslash LC}} & \multicolumn{5}{c||}{\textbf{ORB-SLAM w\textbackslash{}o LC}} & \multicolumn{3}{c|}{\textbf{VITW}} & \multicolumn{3}{c|}{\textbf{Monodepth 2}} \\ \hline
\textbf{Calibration} & $e_f$ & $e_s$ & $m$ & $e_f$/$m$ & $e_s$/$m$ & $e_f$ & $e_s$ & $m$ & $e_f$/$m$ & $e_s$/$m$ & $e$ & $m$ & $e$/$m$ & $e$ & $m$ & $e$/$m$ \\ \hline 
\textbf{Colmap} & 1.01 & {\ul 0.32} & 4.2 & 0.24 & {\ul 0.09} & 2.05 & \textbf{0.28} & 4.1 & 0.58 & \textbf{0.07} & 5.51 & 3.3 & 1.98 & 2.97 & 3.6 & 1.64 \\ 
\textbf{VITW (m)} & 7.94 & 6.19 & 2.3 & 2.74 & 1.67 & 10.03 & 2.85 & 1.8 & 5.71 & 1.94 & 7.07 & 3.4 & 3.49 & 4.17 & 3.6 & 2.61 \\ 
\textbf{VITW turns (m)} & 5.21 & 2.70 & 3.4 & 1.29 & 0.72 & 4.09 & 0.67 & 3.4 & 1.00 & 0.22 & 5.93 & 3.3 & 2.12 & 3.53 & 3.5 & 1.98 \\ \hline
\end{tabular}
\end{table*}

\begin{table*}[tbhp]
\caption{Relative and Absolute 3D Traffic Sign Positioning Error in meters using proposed framework.}
\label{tab:rel_comp}
\centering
\begin{tabular}{|l|cccccccccc|c|}
\hline
\textbf{Seq}        & \textbf{0} & \textbf{1} & \textbf{2} & \textbf{4} & \textbf{5} & \textbf{6} & \textbf{7} & \textbf{8} & \textbf{9} & \textbf{10} & \textbf{Average} \\
\hline
\textbf{Rel} & 0.35 & 1.09 & 0.24 & 2.1 & 0.07 & 0.48 & 0.22 & 0.82 & 0.32 & 0.10 & 0.58 \\
\textbf{Abs} & 0.79 & 1.56 & 0.84 & 4.62 & 0.20 & 1.19 & 0.34 & 4.32 & 0.92 & 0.60 & 1.54 \\
\hline
\end{tabular}%
\end{table*}

Thus, we conclude that for self-calibration, Colmap, VITW (m), and VITW Turns (m) are the better choices. For sign positioning with Approach A using ego-motion estimation, ORB-SLAM (w/ and w/o LC) are the better choices. However, for sign positioning with Approach B using depth estimation, both Monodepth 2 and VITW need to be considered. Finally, it is hypothesized that a combination of multi-view geometry and deep learning approaches is needed for successful sign positioning in all sequences. 

\subsection{3D Positioning Analysis}
\label{sec:3d_analysis}
We compare the accuracy of 3D traffic sign positioning using Approach A against Approach B. We also compare the effect of multi-view geometry and deep learning based self-calibration on the 3D sign positioning accuracy.
We compute the average relative sign positioning error normalized by the number of signs successfully positioned as the metric. 

Table \ref{tab:kitti_rel_final} shows the comparison of the mean performance of 3D traffic sign positioning for the different combinations of self-calibration and depth/ego-motion estimation techniques. The average relative sign positioning error using the full and short trajectories is denoted by $e_f$ and $e_s$ respectively, while $m$ denotes the average number of successfully positioned signs. The relative sign positioning error using depth maps in Approach B is denoted by $e$.
Note that the best average performance is given by Approach A using Colmap for self-calibration and short ORB-SLAM (w/o LC) for ego-motion estimation. The better performance of ORB-SLAM w/o LC for relative sign positioning is explained by the lower ATE-5 (Table \ref{table:ego_motion}) as compared to ORB-SLAM w/ LC. Therefore, approach A using short sub-sequences for triangulation generally performs better than approach B.
However, it is not the case for all the sequences. For Seq 1, where ORB-SLAM fails tracking, the best sign positioning error is given by Approach B using a combination of Colmap for self-calibration and Monodepth 2 for depth estimation. For Seq 4 which does not contain any turns, calibration with Colmap or VITW Turns (m) is not feasible, and VITW (m) has to be used.

While ORB-SLAM (short) w/o LC gives better relative positioning error than w/ LC, the average absolute positioning error is lower when using ORB-SLAM w/ LC (\SI[detect-weight=true, detect-family=true, mode=text]{1.15}{\m}) than w/o LC (\SI[detect-weight=true, detect-family=true, mode=text]{2.46}{\m}). This is because the loop closures help in correcting the accumulated trajectory drift, thereby improving the absolute positions of the traffic signs.

\begin{figure}[tbh!]
\begin{center}
    \includegraphics[width=0.7\linewidth]{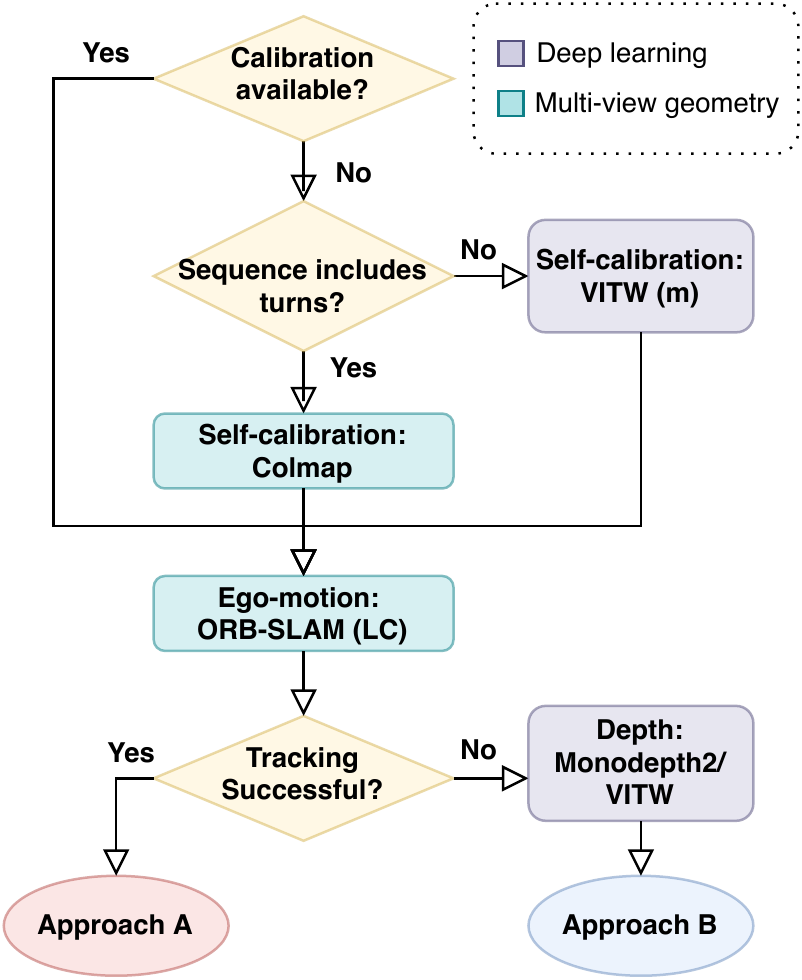}
\end{center}
   \caption{3D Traffic Sign Positioning Framework combining strengths of Multi-View Geometry and Deep Learning for increasing map coverage.}
\label{fig:3D_positioning_flowchart}
\end{figure}

We therefore propose a scheme for crowdsourced 3D traffic sign positioning that combines the strengths of multi-view geometry and deep learning techniques for self-calibration, ego-motion and depth estimation to increase the map coverage. This scheme is shown in Fig. \ref{fig:3D_positioning_flowchart}. The mean relative and absolute 3D traffic sign positioning errors for each validation sequence, computed using this scheme are shown in Table \ref{tab:rel_comp}.
With this approach, our single journey average relative and absolute sign positioning error per sequence is \SI[detect-weight=true, detect-family=true, mode=text]{0.58}{\m} and \SI[detect-weight=true, detect-family=true, mode=text]{1.54}{\m} respectively. The  average  relative positioning  error for all frames is  \SI[detect-weight=true, detect-family=true, mode=text]{0.39}{\m},  while the absolute positioning error for all signs is \SI[detect-weight=true, detect-family=true, mode=text]{1.26}{\m}.
Our relative positioning accuracy is comparable to \cite{dabeer2017end} which unlike our framework uses a camera with known intrinsics, GPS, as well as an IMU to estimate the traffic sign positions. Our absolute positioning accuracy is comparable to \cite{welzel2014accurate}, which also assumes prior knowledge of camera intrinsics as well as the size and height of traffic signs.

%% file: conclusion.tex
\section{Conclusion}
In this paper, we proposed a framework for 3D traffic sign positioning using crowdsourced data from only monocular color cameras and GPS, without prior knowledge of camera intrinsics. We demonstrated that combining the strengths of multi-view geometry and deep learning based approaches to self-calibration, depth and ego-motion estimation results in an increased map coverage. We validated our framework on traffic signs in the public KITTI dataset for single journey sign positioning. 
In the future, the sign positioning accuracy can be further improved with optimization for multiple journeys over the same path.
We will also explore the effect of camera distortion and rolling shutter in the crowdsourced data to expand the scope of our method.

%% file: main.bbl
\begin{thebibliography}{10}
\providecommand{\url}[1]{#1}
\csname url@rmstyle\endcsname
\providecommand{\newblock}{\relax}
\providecommand{\bibinfo}[2]{#2}
\providecommand\BIBentrySTDinterwordspacing{\spaceskip=0pt\relax}
\providecommand\BIBentryALTinterwordstretchfactor{4}
\providecommand\BIBentryALTinterwordspacing{\spaceskip=\fontdimen2\font plus
\BIBentryALTinterwordstretchfactor\fontdimen3\font minus
  \fontdimen4\font\relax}
\providecommand\BIBforeignlanguage[2]{{%
\expandafter\ifx\csname l@#1\endcsname\relax
\typeout{** WARNING: IEEEtran.bst: No hyphenation pattern has been}%
\typeout{** loaded for the language `#1'. Using the pattern for}%
\typeout{** the default language instead.}%
\else
\language=\csname l@#1\endcsname
\fi
#2}}

\bibitem{schwarting2018planning}
W.~Schwarting, J.~Alonso-Mora, and D.~Rus, ``Planning and decision-making for
  autonomous vehicles,'' \emph{Annual Review of Control, Robotics, and
  Autonomous Systems}, vol.~1, no.~1, pp. 187--210, 2018.

\bibitem{jiao2018machine}
J.~{Jiao}, ``Machine learning assisted high-definition map creation,'' in
  \emph{2018 IEEE 42nd Annual Computer Software and Applications Conference
  (COMPSAC)}, vol.~01, 2018, pp. 367--373.

\bibitem{bogdan2018deepcalib}
O.~Bogdan, V.~Eckstein, F.~Rameau, and J.-C. Bazin, ``Deepcalib: A deep
  learning approach for automatic intrinsic calibration of wide field-of-view
  cameras,'' in \emph{Proceedings of the 15th ACM SIGGRAPH European Conference
  on Visual Media Production}, 2018.

\bibitem{gordon2019depthwild}
A.~Gordon, H.~Li, R.~Jonschkowski, and A.~Angelova, ``Depth from videos in the
  wild: Unsupervised monocular depth learning from unknown cameras,'' in
  \emph{The IEEE International Conference on Computer Vision (ICCV)}, 2019.

\bibitem{schonberger2016structure}
J.~L. Schönberger and J.~Frahm, ``Structure-from-motion revisited,'' in
  \emph{Proceedings of the IEEE Conference on Computer Vision and Pattern
  Recognition (CVPR)}, 2016, pp. 4104--4113.

\bibitem{mur2015orb}
R.~Mur-Artal, J.~M.~M. Montiel, and J.~D. Tardos, ``Orb-slam: a versatile and
  accurate monocular slam system,'' \emph{IEEE transactions on robotics},
  vol.~31, no.~5, pp. 1147--1163, 2015.

\bibitem{engel2017direct}
J.~Engel, V.~Koltun, and D.~Cremers, ``Direct sparse odometry,'' \emph{IEEE
  transactions on pattern analysis and machine intelligence}, vol.~40, no.~3,
  pp. 611--625, 2017.

\bibitem{zhou2017unsupervised}
T.~Zhou, M.~Brown, N.~Snavely, and D.~G. Lowe, ``Unsupervised learning of depth
  and ego-motion from video,'' in \emph{Proceedings of the IEEE Conference on
  Computer Vision and Pattern Recognition (ICCV)}, 2017, pp. 1851--1858.

\bibitem{dabeer2017end}
O.~Dabeer, W.~Ding, R.~Gowaiker, S.~K. Grzechnik, M.~J. Lakshman, S.~Lee,
  G.~Reitmayr, A.~Sharma, K.~Somasundaram, R.~T. Sukhavasi, \emph{et~al.}, ``An
  end-to-end system for crowdsourced 3d maps for autonomous vehicles: The
  mapping component,'' in \emph{2017 IEEE/RSJ International Conference on
  Intelligent Robots and Systems (IROS)}.\hskip 1em plus 0.5em minus
  0.4em\relax IEEE, 2017, pp. 634--641.

\bibitem{zhou2018deeptam}
H.~Zhou, B.~Ummenhofer, and T.~Brox, ``Deeptam: Deep tracking and mapping,'' in
  \emph{Proceedings of the European Conference on Computer Vision (ECCV)},
  2018, pp. 822--838.

\bibitem{godard2018digging}
C.~Godard, O.~Mac~Aodha, M.~Firman, and G.~J. Brostow, ``Digging into
  self-supervised monocular depth estimation,'' in \emph{Proceedings of the
  IEEE International Conference on Computer Vision (ICCV)}, 2019, pp.
  3828--3838.

\bibitem{arnoul1996traffic}
P.~Arnoul, M.~Viala, J.~Guerin, and M.~Mergy, ``Traffic signs localisation for
  highways inventory from a video camera on board a moving collection van,'' in
  \emph{Proceedings of Conference on Intelligent Vehicles}.\hskip 1em plus
  0.5em minus 0.4em\relax IEEE, 1996, pp. 141--146.

\bibitem{madeira2005automatic}
S.~Madeira, L.~Bastos, A.~Sousa, J.~Sobral, and L.~Santos, ``Automatic traffic
  signs inventory using a mobile mapping system,'' in \emph{Proceedings of the
  International Conference and Exhibition on Geographic Information GIS
  PLANET}, 2005.

\bibitem{krsak2011traffic}
E.~Krs{\'a}k and S.~Toth, ``Traffic sign recognition and localization for
  databases of traffic signs,'' \emph{Acta Electrotechnica et Informatica},
  vol.~11, no.~4, p.~31, 2011.

\bibitem{welzel2014accurate}
A.~Welzel, A.~Auerswald, and G.~Wanielik, ``Accurate camera-based traffic sign
  localization,'' in \emph{17th International IEEE Conference on Intelligent
  Transportation Systems (ITSC)}.\hskip 1em plus 0.5em minus 0.4em\relax IEEE,
  2014, pp. 445--450.

\bibitem{fairfield2011traffic}
N.~Fairfield and C.~Urmson, ``Traffic light mapping and detection,'' in
  \emph{2011 IEEE International Conference on Robotics and Automation (ICRA)},
  2011, pp. 5421--5426.

\bibitem{chen2016monocular}
X.~Chen, K.~Kundu, Z.~Zhang, H.~Ma, S.~Fidler, and R.~Urtasun, ``Monocular 3d
  object detection for autonomous driving,'' in \emph{Proceedings of the IEEE
  Conference on Computer Vision and Pattern Recognition (CVPR)}, 2016, pp.
  2147--2156.

\bibitem{ku2019monocular}
J.~Ku, A.~D. Pon, and S.~L. Waslander, ``Monocular 3d object detection
  leveraging accurate proposals and shape reconstruction,'' in
  \emph{Proceedings of the IEEE Conference on Computer Vision and Pattern
  Recognition (CVPR)}, 2019, pp. 11\,867--11\,876.

\bibitem{qin2019monogrnet}
Z.~Qin, J.~Wang, and Y.~Lu, ``Monogrnet: A geometric reasoning network for
  monocular 3d object localization,'' in \emph{Proceedings of the AAAI
  Conference on Artificial Intelligence}, vol.~33, 2019, pp. 8851--8858.

\bibitem{zhu2019learning}
J.~Zhu and Y.~Fang, ``Learning object-specific distance from a monocular
  image,'' in \emph{Proceedings of the IEEE International Conference on
  Computer Vision (ICCV)}, 2019, pp. 3839--3848.

\bibitem{bocquillon2007constant}
B.~Bocquillon, A.~Bartoli, P.~Gurdjos, and A.~Crouzil, ``On constant focal
  length self-calibration from multiple views,'' in \emph{2007 IEEE Conference
  on Computer Vision and Pattern Recognition}, 2007, pp. 1--8.

\bibitem{gherardi2010practical}
R.~Gherardi and A.~Fusiello, ``Practical autocalibration,'' in \emph{European
  Conference on Computer Vision (ECCV)}.\hskip 1em plus 0.5em minus 0.4em\relax
  Springer, 2010, pp. 790--801.

\bibitem{de1998self}
L.~de~Agapito, E.~Hayman, and I.~D. Reid, ``Self-calibration of a rotating
  camera with varying intrinsic parameters.'' in \emph{British Machine Vision
  Conference (BMVC)}.\hskip 1em plus 0.5em minus 0.4em\relax Citeseer, 1998,
  pp. 1--10.

\bibitem{pollefeys2008detailed}
M.~Pollefeys, D.~Nist{\'e}r, J.-M. Frahm, \emph{et~al.}, ``Detailed real-time
  urban 3d reconstruction from video,'' \emph{International Journal of Computer
  Vision}, vol.~78, no. 2-3, pp. 143--167, 2008.

\bibitem{lopez2019deep}
M.~Lopez, R.~Mari, P.~Gargallo, Y.~Kuang, J.~Gonzalez-Jimenez, and G.~Haro,
  ``Deep single image camera calibration with radial distortion,'' in
  \emph{Proceedings of the IEEE Conference on Computer Vision and Pattern
  Recognition (CVPR)}, 2019, pp. 11\,817--11\,825.

\bibitem{rong2016radial}
J.~Rong, S.~Huang, Z.~Shang, and X.~Ying, ``Radial lens distortion correction
  using convolutional neural networks trained with synthesized images,'' in
  \emph{Asian Conference on Computer Vision (ACCV)}.\hskip 1em plus 0.5em minus
  0.4em\relax Springer, 2016, pp. 35--49.

\bibitem{workman2015deepfocal}
S.~Workman, C.~Greenwell, M.~Zhai, R.~Baltenberger, and N.~Jacobs, ``Deepfocal:
  A method for direct focal length estimation,'' in \emph{2015 IEEE
  International Conference on Image Processing (ICIP)}.\hskip 1em plus 0.5em
  minus 0.4em\relax IEEE, 2015, pp. 1369--1373.

\bibitem{zhuang2019degeneracy}
B.~Zhuang, Q.-H. Tran, P.~Ji, G.~H. Lee, L.~F. Cheong, and M.~K. Chandraker,
  ``Degeneracy in self-calibration revisited and a deep learning solution for
  uncalibrated slam,'' \emph{2019 IEEE/RSJ International Conference on
  Intelligent Robots and Systems (IROS)}, pp. 3766--3773, 2019.

\bibitem{klein2007parallel}
G.~Klein and D.~Murray, ``Parallel tracking and mapping for small ar
  workspaces,'' in \emph{Proceedings of the 2007 6th IEEE and ACM International
  Symposium on Mixed and Augmented Reality}, 2007, pp. 1--10.

\bibitem{engel2014lsd}
J.~Engel, T.~Sch{\"o}ps, and D.~Cremers, ``Lsd-slam: Large-scale direct
  monocular slam,'' in \emph{European Conference on Computer Vision
  (ECCV)}.\hskip 1em plus 0.5em minus 0.4em\relax Springer, 2014, pp. 834--849.

\bibitem{newcombe2011dtam}
R.~A. Newcombe, S.~J. Lovegrove, and A.~J. Davison, ``Dtam: Dense tracking and
  mapping in real-time,'' in \emph{2011 International Conference on Computer
  Vision (ICCV)}, 2011, pp. 2320--2327.

\bibitem{cao2017estimating}
Y.~Cao, Z.~Wu, and C.~Shen, ``Estimating depth from monocular images as
  classification using deep fully convolutional residual networks,'' \emph{IEEE
  Transactions on Circuits and Systems for Video Technology}, vol.~28, no.~11,
  pp. 3174--3182, 2017.

\bibitem{eigen2014depth}
D.~Eigen, C.~Puhrsch, and R.~Fergus, ``Depth map prediction from a single image
  using a multi-scale deep network,'' in \emph{Advances in Neural Information
  Processing Systems (NeurIPS)}, 2014, pp. 2366--2374.

\bibitem{liu2015learning}
F.~Liu, C.~Shen, G.~Lin, and I.~Reid, ``Learning depth from single monocular
  images using deep convolutional neural fields,'' \emph{IEEE transactions on
  pattern analysis and machine intelligence}, vol.~38, no.~10, pp. 2024--2039,
  2015.

\bibitem{wang2017deepvo}
S.~Wang, R.~Clark, H.~Wen, and N.~Trigoni, ``Deepvo: Towards end-to-end visual
  odometry with deep recurrent convolutional neural networks,'' in \emph{2017
  IEEE International Conference on Robotics and Automation (ICRA)}.\hskip 1em
  plus 0.5em minus 0.4em\relax IEEE, 2017, pp. 2043--2050.

\bibitem{casser2019unsupervised1}
V.~Casser, S.~Pirk, R.~Mahjourian, and A.~Angelova, ``Unsupervised learning of
  depth and ego-motion: A structured approach,'' in \emph{AAAI Conference on
  Artificial Intelligence (AAAI)}, vol.~2, 2019, p.~7.

\bibitem{godard2017unsupervised}
C.~Godard, O.~Mac~Aodha, and G.~J. Brostow, ``Unsupervised monocular depth
  estimation with left-right consistency,'' in \emph{Proceedings of the IEEE
  Conference on Computer Vision and Pattern Recognition (ICCV)}, 2017, pp.
  270--279.

\bibitem{li2018undeepvo}
R.~Li, S.~Wang, Z.~Long, and D.~Gu, ``Undeepvo: Monocular visual odometry
  through unsupervised deep learning,'' in \emph{2018 IEEE International
  Conference on Robotics and Automation (ICRA)}.\hskip 1em plus 0.5em minus
  0.4em\relax IEEE, 2018, pp. 7286--7291.

\bibitem{zhan2018unsupervised}
H.~Zhan, R.~Garg, C.~Saroj~Weerasekera, K.~Li, H.~Agarwal, and I.~Reid,
  ``Unsupervised learning of monocular depth estimation and visual odometry
  with deep feature reconstruction,'' in \emph{Proceedings of the IEEE
  Conference on Computer Vision and Pattern Recognition (CVPR)}, 2018, pp.
  340--349.

\bibitem{yang2018deep}
N.~Yang, R.~Wang, J.~Stuckler, and D.~Cremers, ``Deep virtual stereo odometry:
  Leveraging deep depth prediction for monocular direct sparse odometry,'' in
  \emph{Proceedings of the European Conference on Computer Vision (ECCV)},
  2018, pp. 817--833.

\bibitem{ramer1972iterative}
U.~Ramer, ``An iterative procedure for the polygonal approximation of plane
  curves,'' \emph{Computer graphics and image processing}, vol.~1, no.~3, pp.
  244--256, 1972.

\bibitem{douglas1973algorithms}
D.~H. Douglas and T.~K. Peucker, ``Algorithms for the reduction of the number
  of points required to represent a digitized line or its caricature,''
  \emph{Cartographica: the international journal for geographic information and
  geovisualization}, vol.~10, no.~2, pp. 112--122, 1973.

\bibitem{umeyama1991least}
S.~{Umeyama}, ``Least-squares estimation of transformation parameters between
  two point patterns,'' \emph{IEEE Transactions on Pattern Analysis and Machine
  Intelligence}, vol.~13, no.~4, pp. 376--380, 1991.

\bibitem{szeliski2010computer}
R.~Szeliski, \emph{Computer vision: algorithms and applications}.\hskip 1em
  plus 0.5em minus 0.4em\relax Springer Science \& Business Media, 2010.

\bibitem{geiger2012we}
A.~Geiger, P.~Lenz, and R.~Urtasun, ``Are we ready for autonomous driving? the
  kitti vision benchmark suite,'' in \emph{IEEE Conference on Computer Vision
  and Pattern Recognition}, 2012, pp. 3354--3361.

\bibitem{horn1987closed}
B.~K.~P. Horn, ``Closed-form solution of absolute orientation using unit
  quaternions,'' \emph{Journal of the Optical Society of America A}, vol.~4,
  no.~4, pp. 629--642, 1987.

\end{thebibliography}
